\documentclass{article}

\usepackage{PRIMEarxiv}

\usepackage[utf8]{inputenc} 
\usepackage[T1]{fontenc}    
\usepackage{hyperref}       
\usepackage{url}            
\usepackage{booktabs}       
\usepackage{amsfonts}       
\usepackage{nicefrac}       
\usepackage{microtype}      
\usepackage{lipsum}
\usepackage{fancyhdr}       
\usepackage{graphicx}       
\usepackage{algorithm}
\usepackage{algorithmic}
\usepackage{amsmath}
\usepackage{amssymb}
\usepackage{newfloat}
\usepackage{listings}
\usepackage{array}
\usepackage{multirow}
\usepackage{pifont}
\usepackage{caption} 
\usepackage{color} 
\usepackage{enumitem}
\usepackage[square,sort,comma,numbers]{natbib}
\usepackage{xcolor}
\usepackage{subcaption} 

\definecolor{darkgreen}{rgb}{0.0, 0.5, 0.0}
\DeclareCaptionStyle{ruled}{labelfont=normalfont,labelsep=colon,strut=off}

\floatstyle{ruled}
\newfloat{listing}{tb}{lst}{}
\floatname{listing}{Listing}

\graphicspath{{media/}}     

\title{Multimodal Generalized Category Discovery
}

\author{
  \textbf{Yuchang Su}$^{1}$ \quad
  \textbf{Renping Zhou}$^{1}$ \quad
  \textbf{Siyu Huang}$^{2}$ \quad
  \textbf{Xingjian Li}$^{3}$ \quad \\
  \textbf{Tianyang Wang}$^{1}$ \quad 
  \textbf{Ziyue Wang}$^{3}$ \quad
  \textbf{Min Xu}$^{3}\thanks{Corresponding author.}$ \\[0.5em]
  $^{1}$University of Alabama at Birmingham \quad
  $^{2}$Clemson University \quad
  $^{3}$Carnegie Mellon University
   \\[0.5em]
  \texttt{\{syccc142857,tehaji007,lixj04\}@gmail.com},  \texttt{siyuh@clemson.edu}, \\
  \texttt{tw2@uab.edu}, 
  \texttt{ziyuew2@andrew.cmu.edu}, 
  \texttt{mxu1@cs.cmu.edu} 
}

\begin{document}
\maketitle

\begin{abstract}
Generalized Category Discovery (GCD) aims to classify inputs into both known and novel categories, a task crucial for open-world scientific discoveries. However, current GCD methods are limited to unimodal data, overlooking the inherently multimodal nature of most real-world data. In this work, we extend GCD to a multimodal setting, where inputs from different modalities provide richer and complementary information. Through theoretical analysis and empirical validation, we identify that the key challenge in multimodal GCD lies in effectively aligning heterogeneous information across modalities. To address this, we propose MM-GCD, a novel framework that aligns both the feature and output spaces of different modalities using contrastive learning and distillation techniques. MM-GCD achieves new state-of-the-art performance on the UPMC-Food101 and N24News datasets, surpassing previous methods by 11.5\% and 4.7\%, respectively.

\end{abstract}


\section{Introduction}

Generalized Category Discovery (GCD) \cite{Vaze_2022_CVPR} aims to classify new inputs into both known and unknown classes, making it particularly valuable for open-world scientific discoveries. 
For instance, GCD has the potential to analyze genetic data from patients and discover new variants associated with rare diseases, revealing novel disease categories that extend beyond known classifications~\cite{10.1093/ckj/sfx051}. This capability is of great interest in rare disease research.

However, existing research \cite{wang2024sptnet, an2023generalizedcategorydiscoverydecoupled} has predominantly focused on unimodal data, neglecting the inherently multimodal nature of real-world scenarios. For example, radiology images (e.g., X-rays, CT scans) are frequently paired with clinical reports, each contributing unique information. Leveraging this rich, multimodal data enhances decision-making accuracy, similar to how radiologists use both images and clinical reports for more accurate diagnoses.

\begin{figure}[!tb]
    \centering
    \hspace*{-0.035\columnwidth} %
    \resizebox{0.55\columnwidth}{!}{%
        \includegraphics[width=2.1\columnwidth]{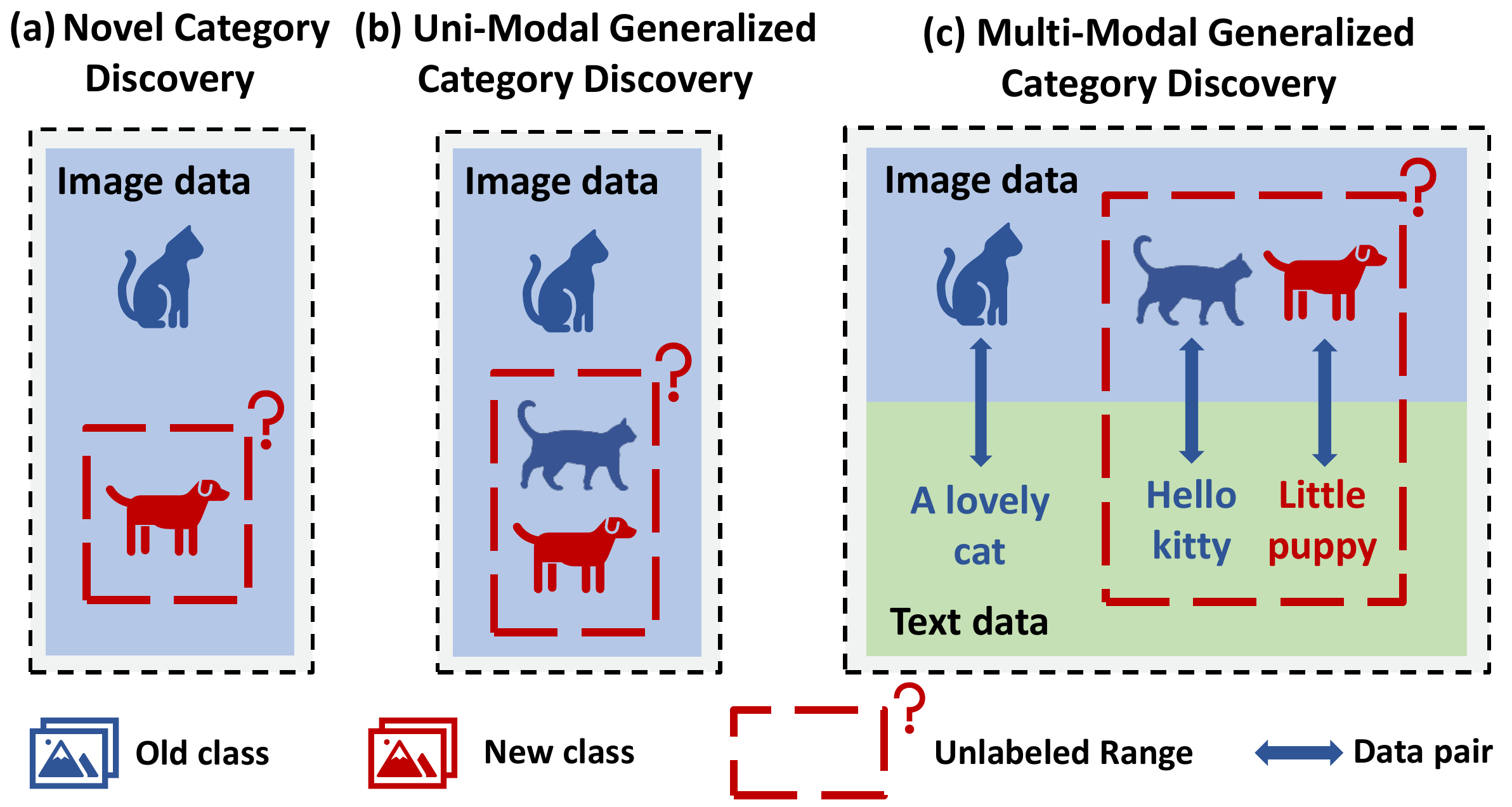}%
    }%
    \vspace{-0.5em}
    \caption{Evolution from NCD to Multimodal GCD.  (\textbf{a}) Novel Category Discovery (NCD) dealt with unlabelled images containing only new classes. (\textbf{b}) Generalized Category Discovery (GCD) expanded this by including possible old classes in the unlabelled set but was limited to single modality data. (\textbf{c}) Our multimodal-GCD model addresses these limitations by focusing on multimodal data which is now abundantly present in real life, and leveraging inter-modality interactions to improve learning where labels are missing.}
    \label{fig:task}
    \vspace{-0.5em}
\end{figure}

In this work, we extend GCD to the multi-modal setting, where inputs from different modalities are simultaneously used for the classification process (Figure \ref{fig:task}). Multimodal GCD presents unique challenges compared to its unimodal counterpart: how can we effectively utilize the rich yet heterogeneous information from different modalities? We find that simply extending existing GCD frameworks to multimodal settings is subject to degraded performance, as the information across modalities is often poorly aligned, making the model biased to spurious correlations and irrelevant information. In \S\ref{sec:theory}, we provide theoretical analysis and empirical validation showing that better-aligned multimodal data reduces variance and simplifies the decision boundary, improving classification performance.

Building on these insights, we propose a novel framework, MM-GCD, that directly addresses the alignment challenge in multimodal GCD. We divide the GCD learning process into two stages: the first is generating the feature space, and the second is partitioning it, which corresponds to the results in the output space. Theoretical analysis has already proven that alignment is indispensable in the feature space. We resort to recent advances in multimodal contrastive learning \cite{pmlr-v139-radford21a} to ensure that features from different modalities are properly aligned. The goal of multimodal contrastive learning is to bring similar concepts from different modalities closer while pushing dissimilar concepts apart. This alignment results in similar concepts from different modalities having similar features, making them easier for models to understand. Additionally, given the classification nature of multimodal GCD, in the output space, we leverage distillation techniques to ensure that classification results are consistent across modalities. Distillation allows predictions from one modality to serve as targets for the other, combined with entropy minimization to ensure consistency between modalities. By aligning both the feature and output spaces, MM-GCD effectively integrates the heterogeneous information from different modalities, providing a effective solution to the GCD task.

We validate the effectiveness of MM-GCD on two benchmarks and perform ablation studies on its components. MM-GCD sets a new state-of-the-art on the UPMC-Food-101 \cite{wang2015recipe} and N24News \cite{wang-etal-2022-n24news} datasets, surpassing previous methods by 11.5\% and 4.7\%, respectively. Leveraging both modalities leads to improvements of 6.8\% and 3.4\% over single-modality approaches, highlighting that different modalities provide complementary and enriched information for the GCD task. Visualizations of the feature space confirm that our alignment objectives effectively close the modality gap and synchronize the feature spaces. Ablation studies show that, without these alignments, performance drops by 18.8\% on Food101, falling below even the single-modality baseline.

In summary, our work makes the following contributions:

\setlength{\itemindent}{0pt}
\begin{itemize}
\item We introduce the multimodal Generalized Category Discovery setting, which closely mirrors real-world scenarios where data naturally exist in multiple modalities.
\item We theoretically and empirically demonstrate that modality alignment is the most critical aspect for successfully tackling the multimodal Generalized Category Discovery problem.
\item We propose novel and effective alignment methods that solve this problem, achieving new state-of-the-art performance on the Food101 and N24News datasets, with improvements of 11.5\% and 4.7\%, respectively.
\end{itemize}

\section{Related Works}

\textbf{Multimodal Learning} aims to integrate features across multiple modalities, such as audio, text, and images, to enhance task performance \cite{inproceedings, 8269806}. The primary focus in this field is on fusion strategies, categorized into aggregation-based and alignment-based methods \cite{8269806, zou2023unismmc}. Aggregation methods, like feature concatenation, are popular \cite{nojavanasghari, anastasopoulos2019neural, wang-etal-2022-n24news}, while alignment-based methods, such as contrastive learning, align and fuse modalities by pre-training on large datasets \cite{pmlr-v139-radford21a, yuan2021multimodal}. These approaches leverage complementary strengths from different modalities, improving classification accuracy and robustness \cite{8269806}. 

\textbf{Novel Class Discovery (NCD)} tackles the identification and categorization of new, unseen classes within unlabeled datasets. Early work, such as \cite{han2020automatically}, introduced deep clustering algorithms that significantly improved novel class discovery in image datasets. These methods inspired further research, extending to multimodal settings with cross-modal discrimination and hierarchical frameworks \cite{jia2021joint, li2023reinforcement}. Despite these advances, NCD methods are not designed for GCD tasks, which require classification of both knowns and unknowns. 

\textbf{Generalized Category Discovery (GCD)} extends NCD by requiring the classification of both seen and unseen categories within unlabeled data \cite{Vaze_2022_CVPR}. Methods like SimGCD \cite{Wen_2023_ICCV}, PromptCAL \cite{zhang2023promptcal}, DPN \cite{an2023generalizedcategorydiscoverydecoupled} and SPTNet \cite{wang2024sptnet} have advanced GCD by addressing biases towards known classes and improving learning phase interactions. However, these approaches are primarily unimodal, focusing on either visual or textual data, and do not leverage the full potential of multimodal information. Unlike these methods, our approach optimizes existing paired multimodal data by capturing shared information and reducing noise across modalities, leading to more robust GCD performance.

Current models are often constrained by their reliance on unimodal information, typically employing strategies designed to generate multimodal content from unimodal inputs. For example, GET \cite{wang2024get} converts visual embeddings into text tokens for use in CLIP's text encoder, while CLIP-GCD \cite{ouldnoughi2023clip} retrieves textual descriptions from large text databases to enhance image understanding. TextGCD \cite{zheng2024textual} generates descriptive text through retrieval-based strategies. While these methods improve unimodal data interpretation, they fail to fully exploit the rich multimodal data. In real-world scenarios, modality correlations are lower, yet the information is richer. Effectively harnessing this diverse data while minimizing noise is crucial but challenging, and our method addresses this by aligning and integrating multimodal information.
\begin{figure*}[!tb]
    \centering
    \resizebox{1.01\textwidth}{!}{%
        \includegraphics[width=0.63\columnwidth]{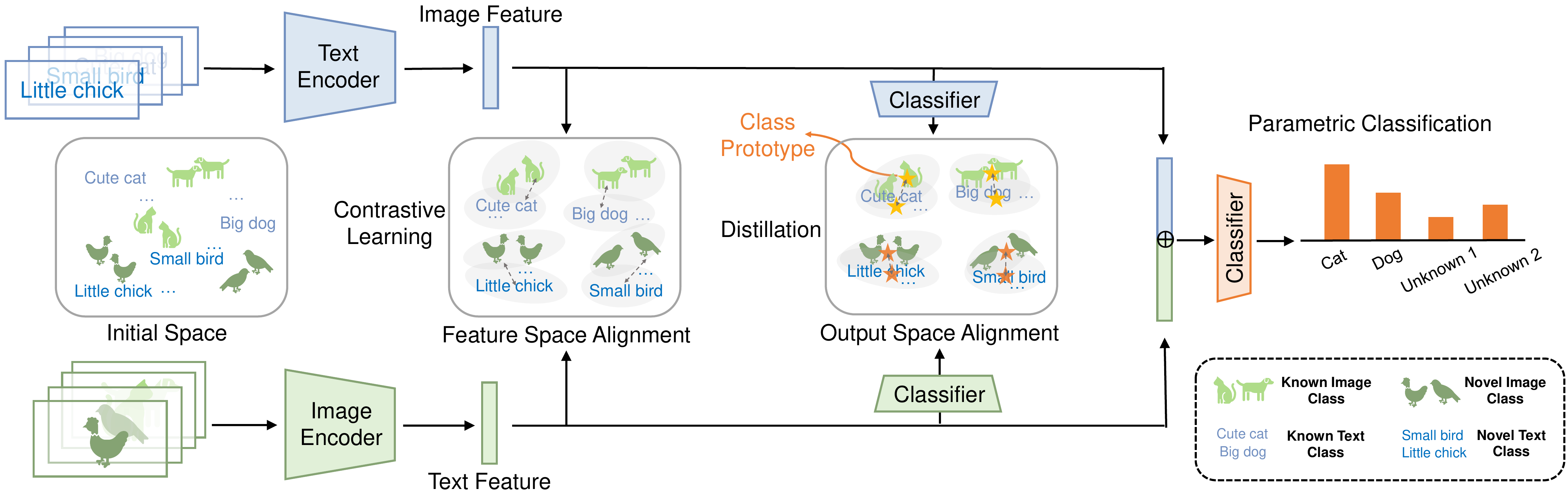}
    }%
    \vspace{-0.5em}
    \caption{Overview of our MM-GCD framework: We propose a dual-branch structure that processes text and image data separately, calculating unimodal loss to capture category distinctions. Our framework focuses on aligning the feature space through multimodal contrastive learning and optimizing the output space by entropy minimization for consistent decision-making across modalities. }
    \label{fig:model}
    \vspace{-0.5em}
\end{figure*}
\section{Problem Formulation}

In this work, we introduce the novel problem of \emph{Multimodal Generalized Category Discovery (Multimodal GCD)}. The goal of Multimodal GCD is to identify both known and unknown categories from multimodal data, where the model has access to labeled data for some categories and must discover new categories in unlabeled data. This problem is crucial for applications where data comes from multiple modalities, such as text and images, and where not all categories are known in advance.

Formally, let \( \mathcal{D}^l = \{(\mathbf{x}_i^l, \mathbf{y}_i^l)\} \subset \mathcal{X} \times \mathcal{Y}_{\text{old}} \) denote the labeled dataset, where each instance \( \mathbf{x}_i^l = (\mathbf{x}_{i1}, \mathbf{x}_{i2}) \) consists of inputs from two different modalities (e.g., text and images), and \( \mathbf{y}_i^l \) is its corresponding label. Here, \( \mathcal{X} = \mathcal{X}_1 \times \mathcal{X}_2 \) represents the multimodal input space, and \( \mathcal{Y}_{\text{old}} \) is the label space containing only known categories.

In addition, we have an unlabeled dataset \( \mathcal{D}^u = \{\mathbf{x}_i^u\} \subset \mathcal{X} \), where \( \mathbf{x}_i^u = (\mathbf{x}_{i1}, \mathbf{x}_{i2}) \) consists of the same two modalities. 
The corresponding labels  \(\mathcal{Y} = \mathcal{Y}_{\text{old}} \cup \mathcal{Y}_{\text{new}}\) include both known categories \( \mathcal{Y}_{\text{old}} \) and unknown categories \( \mathcal{Y}_{\text{new}} \), but these labels are not available to model during training. 

The objective is to learn a model that accurately predicts labels for both known and unknown categories in the unlabeled dataset. The number of unknown categories \(|\mathcal{Y}_{\text{new}}| \) may be given as a prior or estimated during training. The challenge lies in the simultaneous discovery of new categories and alignment of known categories across multiple modalities, which is critical for advancing multimodal understanding in open-world settings.

\section{Theoretical Analysis}
\label{sec:theory}

In our initial attempt to extend Generalized Category Discover (GCD) from unimodal to multimodal, we use CLIP's encoders to extract features separately from text and image modalities, and then concatenating these features to form a new fusion feature. This approach, while seemingly promising, surprisingly resulted in decreased performance compared to using unimodal information alone (Table~\ref{tab:main_result} in \S\ref{sec:results}).

Through detailed analysis, we identified that the primary issue was the misalignment between text and image features. This misalignment led to a dilution of useful information, negatively impacting the model's effectiveness. For example, a picture of a dancer might be paired with textual descriptions that provide relevant but inconsistent information, which, if not properly addressed, could lead to confusion for the model. These findings highlight the importance of ensuring proper alignment between different modalities to fully leverage the benefits of multimodal data. Therefore, strengthening this alignment is crucial to achieving better integration and performance in multimodal tasks.


To further explore this, we conducted theoretical derivations from the perspective of data distribution, providing insights into the impact of modality alignment on the task. For a multimodal dataset \((X, Y)\), where \(X\) and \(Y\) are random variables representing features from two different modalities, we assume that \(X_k\) and \(Y_k\) are the random variables corresponding to the features belonging to the \(k\)-th category. Here, we deliberately abuse the notation \(Y\) to represent features from another modality, rather than the labels as defined in Section \S3, for the sake of clearer representation. When the data volume is large, \(X_k\) and \(Y_k\) are assumed to follow multivariate normal distributions, as shown in Equation \ref{eq:def}:
\begin{equation}
X_k \sim \mathcal{N}\left(\mu_{X_k}, \mathcal{S}_{X_k}\right), \ \ \ 
Y_k \sim \mathcal{N}\left(\mu_{Y_k}, \mathcal{S}_{Y_k}\right) 
\label{eq:def}
\end{equation}
where $\mu_{X_k}, \mu_{Y_k}$ denote the mean values of the corresponding data from the two modalities, while $\mathcal{S}_{X_k}, \mathcal{S}_{Y_k}$ denote the covariance matrices.
We formulate the fused modality $F_k$ with a simple concatenation of the two modalities $X_k, Y_k$):
\begin{equation}
    F_k = X_k \oplus Y_k \sim \mathcal{N}\left(\mu_{F_k}, \mathcal{S}_{F_k}\right) 
    \label{eq:def_F}
\end{equation}
$\mu_{F_k}$ and $\mathcal{S}_{F_k}$  represent the mean and covariance matrix of $F_k$, respectively. The distribution statistics of $F_k$ are:

\begin{equation}
    \mu_{F_k} = \begin{pmatrix}
  \mu_{X_k}  \\
  \mu_{Y_k} 
\end{pmatrix} \ , \
\mathcal{S}_F = \begin{pmatrix}
  \mathcal{S}_{X_k} & \mathcal{S}_{X_kY_k}  \\
  (\mathcal{S}_{X_kY_k})^T & \mathcal{S}_{Y_k}
\end{pmatrix}
\mathcal{S}_{X_kY_k}
\label{eq:fusion-multi-gaussian}
\end{equation}
\begin{equation}
    \mathcal{S}_{X_kY_k} = \mathcal{S}_{X_k}^\frac{1}{2}R_k\mathcal{S}_{Y_k}^\frac{1}{2}
\end{equation}

\begin{equation}
R_k = \begin{pmatrix}
  \sigma_{X_k^{(1)},Y_k^{(1)}} & & & \\ 
  & \sigma_{X_k^{(2)},Y_k^{(2)}}  & & \\ 
  & & \cdots  & \\ 
  & & &\sigma_{X_k^{(n)},Y_k^{(n)}} \\ 
\end{pmatrix}
\end{equation}
$R_k$ is a diagonal matrix composed of the correlation coefficients, where $\sigma_{X_k^{(i)},Y_k^{(i)}} \in \left(0,1\right)$ converges to 1 when the distributions of the two modalities are perfectly correlated.

Here we use the determinant of the covariance matrix of a distribution as the objective function $\mathcal{L}$ to measure the distribution of the data:
\begin{equation}
\mathcal{L}_{X} = \sum_{k \in \mathcal{C}} {\lvert \mathcal{S}_{X_k}\rvert} \ \ , \
\mathcal{L}_{Y} = \sum_{k \in \mathcal{C}} {\lvert \mathcal{S}_{Y_k}\rvert} \ \ , \
\mathcal{L}_{F} = \sum_{k \in \mathcal{C}} {\lvert \mathcal{S}_{F_k}\rvert}
\end{equation}
The determinant can be considered as a measure of volume in high-dimensional space, the smaller this objective function, the more compact the within-class distribution of the data, such that the data is potentially easier to be clustered by clustering algorithms.

We then give the relationship between the objective function of the fused modality and the original distributions of the two modalities, as:
\begin{align}
\mathcal{L}_{F} 
&= \sum_{k \in \mathcal{C}}\lvert\mathcal{S}_{X_k}\rvert \cdot \lvert \mathcal{S}_{Y_k} - \mathcal{S}_{X_kY_k}^T \mathcal{S}_{X_k}^{-1} \mathcal{S}_{X_kY_k}\rvert \\
&= \sum_{k \in \mathcal{C}}\lvert\mathcal{S}_{X_k}\rvert \cdot \lvert\mathcal{S}_{Y_k} - \left(\mathcal{S}_{X_k}^\frac{1}{2}R_k\mathcal{S}_{Y_k}^\frac{1}{2}\right)^T \mathcal{S}_{X_k}^{-1} \left(\mathcal{S}_{X_k}^\frac{1}{2}R_k\mathcal{S}_{Y_k}^\frac{1}{2}\right)\rvert \\
&= \sum_{k \in \mathcal{C}}\lvert\mathcal{I}-R_k^2\rvert \cdot \lvert\mathcal{S}_{X_k}\rvert \cdot \lvert\mathcal{S}_{Y_k}\rvert
\label{eq:align}
\end{align}

From the final result, it can be seen that $\mathcal{L}_F$ is negatively correlated with $R_k$. This means that as $R_k$ becomes larger, and the two distributions become closer, $\mathcal{L}_F$ decreases, increasing the compactness of the joint distribution. This suggests that even if the individual results of the two distributions do not improve, i.e., when $\mathcal{L}_X$ and $\mathcal{L}_Y$ remain unchanged, we can still make the joint distribution more distinguishable by improving the alignment of the modalities.

\begin{figure}[!tb]
  \centering
  \includegraphics[width=0.8\columnwidth, height=0.4\columnwidth]{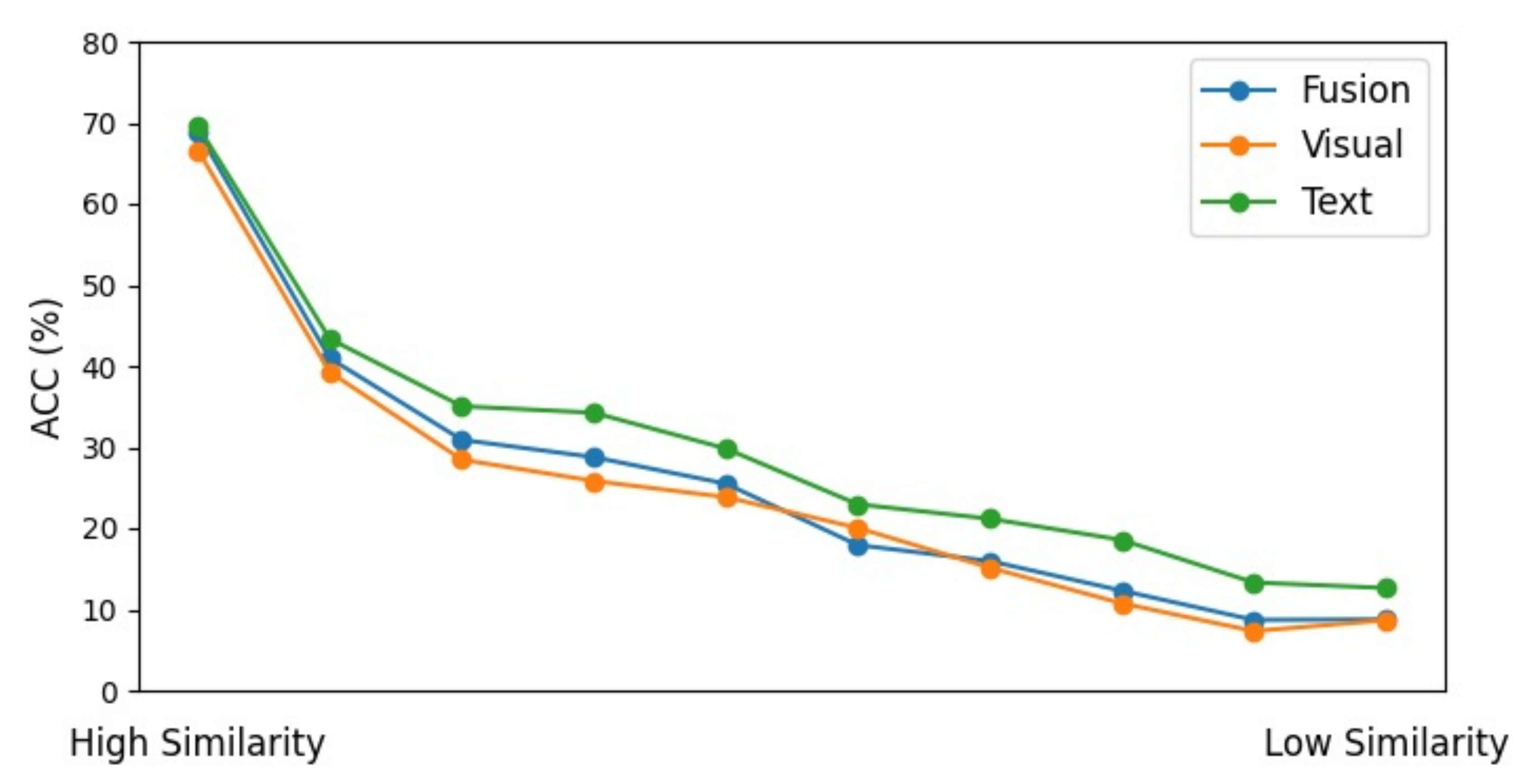}
  \vspace{-0.5em}
  \caption{The relationship between accuracy and feature similarity across visual and text modalities. Result shows that groups with higher feature similarity tend to achieve greater accuracy.}
  \vspace{-0.5em}
  \label{fig:acc-sim}
\end{figure}

We conducted experiments to prove our theory: using the baseline method, we sort the test data based on the cosine similarity of cross-modality features between the text and visual and subsequently divided into ten groups. As shown in Fig. \ref{fig:acc-sim}, the accuracy within each group revealed that groups with higher cross-modality feature similarity exhibited higher accuracy. This finding suggests that there is a strong correlation between the alignment of features across modalities and the resulting accuracy.

\section{Method}

Through theoretical derivation and experimental validation, we have identified that the key to the problem is the alignment between modalities. With this objective in mind, we have reconsidered the goal of the loss function in the uni-modal setting and incorporated the goal of inter-modal alignment into it.

\subsection{What is required to discover a new category?}

The requirements for the Generalized Category Discovery (GCD) problem can be divided into two key aspects: \textit{creating an embedding space that can differentiate each category} and \textit{effectively partitioning this embedding space to include new categories}. These aspects correspond to the goals of representation learning \cite{Vaze_2022_CVPR} and parametric classification \cite{Wen_2023_ICCV} in previous methods. In our approach, we incorporate the objective of modality alignment into the learning processes of these two components. In \S5.2, we focus on aligning the features of different modalities within the embedding space, while in \S5.3, we address the unified partitioning of this space.

Let us define the basic notations used in this section. Given a sample \( \mathbf{x}_i \) from one modality and its augmented view \( \mathbf{x}_i' \), along with the corresponding data \( \mathbf{y}_i \) from another modality within the same data pair, let \( \mathbf{h}_i \) denote the feature representation of \( \mathbf{x}_i \) after passing through the backbone encoder \( f_x \), such that \( \mathbf{h}_i = f_x(\mathbf{x}_i) \). Similarly, let \(\widetilde{\mathbf{h}}_i\) represent the feature obtained from \( \mathbf{y}_i \) after passing through the backbone encoder \( f_y \), such that \(\widetilde{\mathbf{h}}_i = f_y(\mathbf{y}_i) \). Additionally, let \( \mathbf{z}_i \) represent the output of a projection layer \( g \) applied to \( \mathbf{h}_i \), followed by normalization, such that \( \mathbf{z}_i = \text{norm}(g(\mathbf{h}_i)) \).

\subsection{Constructing Embedding Space}

To build a robust embedding space across different modalities, representation learning with contrastive loss is widely employed. A critical aspect of this process is the selection of positive pairs. In the multimodal setting, we define three types of positive pairs.


The first type involves an unsupervised contrastive loss between two views \( \mathbf{x}_i \) and \( \mathbf{x}_i' \) of the same sample within a mini-batch \( B \):
\begin{equation}
\mathcal{L}^{u}_{\text{rep}} = \frac{1}{|B|} \sum_{i \in B} -\log \frac{\exp(\mathbf{z}_i^\top \mathbf{z}_i' / \tau_u)}{\sum_{j \neq i} \exp(\mathbf{z}_i^\top \mathbf{z}_j' / \tau_u)}.
\end{equation}

The second type introduces a supervised contrastive loss between a sample and other samples with the same label within a labeled mini-batch \( B^l \):
\begin{equation}
\mathcal{L}^{s}_{\text{rep}} = \frac{1}{|B^l|} \sum_{i \in B^l} \frac{1}{|\mathcal{N}_i|} \sum_{r \in \mathcal{N}_i} -\log \frac{\exp(\mathbf{z}_i^\top \mathbf{z}_r' / \tau_s)}{\sum_{j \neq i} \exp(\mathbf{z}_i^\top \mathbf{z}_j' / \tau_s)}.
\end{equation}
where \( \mathcal{N}_i \) is the sample set that shares the same label as \( x_i \).

The third type applies a cross-modal contrastive loss, aimed at aligning features across modalities. Here, the positive pairs extend across different modalities within the same data pair, bringing their respective embedding spaces closer. Given that the encoders from CLIP are already trained for alignment, we use the features \( \mathbf{h}_i \) directly from the encoders for this cross-modal contrastive loss, bypassing the additional projection layer:
\begin{equation}
\mathcal{L}_{\text{rep}}^c = \frac{1}{|B|} \sum_{i \in B} -\log \frac{\exp(\mathbf{h}_i^\top \widetilde{\mathbf{h}}_i / \tau_c)}{\sum_{j \neq i} \exp(\mathbf{h}_i^\top \widetilde{\mathbf{h}}_j / \tau_c)}.
\end{equation}

The temperatures \( \tau_u \), \( \tau_s \), and \( \tau_c \) control the scaling in each level of contrastive learning. We balance the different representation losses using coefficients \( \lambda_u \) and \( \lambda_s \). The overall representation learning objective is then defined as:
\begin{equation}
\mathcal{L}_{\text{rep}} = \lambda_u \mathcal{L}^u_{\text{rep}} + \lambda_s \mathcal{L}^s_{\text{rep}} + (1 - \lambda_u - \lambda_s)\mathcal{L}_{\text{rep}}^c.
\end{equation}

\subsection{Partitioning Embedding Space} 

After constructing the embedding space, it is crucial to effectively partition this space for classification, especially when dealing with both known and unknown categories. Previous work \cite{Wen_2023_ICCV} has demonstrated that parametric methods for partitioning can yield superior results. Building on this insight, we develop a trainable prototypical classifier, denoted as $\mathcal{C} = \{\mathbf{c}_1, \dots, \mathbf{c}_K\}$, where $K$ represents the total number of categories, including both old and new classes. For each sample \( \mathbf{x}_i \) in a mini-batch, we compute a pseudo-probability \( p_i \) over all categories:
\begin{equation}
p_i^{(k)} = \frac{\exp(\mathbf{h}_i^\top \mathbf{c}_k / \tau_p)}{\sum_{k'} \exp(\mathbf{h}_i^\top \mathbf{c}_{k'} / \tau_p)}.
\end{equation}






Similar to the representation learning section, we also present three levels of classification loss.
The first loss is self-distillation of all samples. We produce 
an additional soft pseudo-label \( \mathbf{q}_i \) 
that
is generated from a different view of \( \mathbf{x}_i \) using a more acute temperature \( \tau_q \). We minimize the cross-entropy loss \( \ell(\mathbf{q}, \mathbf{p}) = - \sum_k \mathbf{q}^{(k)} \log \mathbf{p}^{(k)} \) between the predicted probabilities and the pseudo-labels for all samples in a mini-batch $B$:
\begin{equation}
\mathcal{L}^u_{\text{cls}} = \frac{1}{|B|} \sum_{i \in B} \ell(\mathbf{q}_i, \mathbf{p}_i)
\end{equation}

The second is towards all the labeled set $B_l$. Using this existing supervision, we can quickly correct the positions of some old class prototypes, thereby creating a clearer partition of the entire space:
\begin{equation}
\mathcal{L}^u_{\text{cls}} = \frac{1}{|B|} \sum_{i \in B} \ell(\mathbf{y}_i, \mathbf{p}_i)
\end{equation}
where \( \mathbf{y}_i \) is the actual label of \( \mathbf{x}_i \).

\begin{table}[!tb]   
    
    \begin{center}
    \resizebox{0.45\textwidth}{!}{%
    \begin{tabular}{lcccc}
    \toprule
             & \multicolumn{2}{c}{Labelled}  & \multicolumn{2}{c}{Unlabelled}\\
                \cmidrule(rl){2-3}\cmidrule(rl){4-5}
    Dataset      & \#Data   & \#Class   & \#Data   & \#Class \\
    \midrule
    UPMC-Food101 \cite{wang2015recipe}   & 15K     & 50         & 45K     & 101 \\
    N24News \cite{wang-etal-2022-n24news} & 12K     & 12        & 36K     & 24 \\
    \bottomrule
    \end{tabular}%
    }
    \end{center}
    \caption{Statistics of the multimodal classification benchmarks used for evaluation. }
    \vspace{-1em}
    \label{tab:split}
\end{table}

The third is multimodal prototype distillation that we propose. It effectively leverages the pairwise information among modalities in a self-distillation manner: each modality distills the classification decisions made by models of other modalities. It ensures that samples in a pair occupy analogous positions in the latent image and text spaces, facilitating the clustering/prototyping process that plays a key role in the the problem:
\begin{equation}
\mathcal{L}^c_{\text{cls}} = \frac{1}{|B|} \sum_{i \in B} \ell(\widetilde{\mathbf{p}_i}, \mathbf{p}_i)
\end{equation}
where  $\widetilde{\mathbf{p}_i}$ is the pseudo probabilities of the corresponding data $y_i$. By strengthening the decision consistency among modalities, the resulting output become more robust to cross-modal distributional discrepancies, thereby enhancing the clustering/prototyping process in multimodal latent spaces. 
Since the classification results for unknown classes from different modalities may be inconsistent (e.g., assigning the same new class to different clusters), we applied the Hungarian algorithm \cite{kuhn1955hungarian} for label mapping to determine the optimal classification correspondence before calculating the entropy loss.

A mean-entropy regularization term $H (\overline{\mathbf{p}})$ is additionally adopted to add balance of the classification result as a learning objective, where \( \overline{\mathbf{p}} = \frac{1}{2|B|} \sum_{i \in B} (\mathbf{p}_i + \mathbf{p}'_i) \) signifies the average prediction for a batch, and the entropy \( H(\overline{\mathbf{p}}) = - \sum_k \overline{\mathbf{p}}^{(k)} \log \overline{\mathbf{p}}^{(k)} \) measures the uncertainty of this prediction. The overall parametric classification objective is:
\begin{equation}
\mathcal{L}_{\text{cls}} = \lambda_u \mathcal{L}^u_{\text{cls}} + \lambda_s \mathcal{L}^s_{\text{cls}} + (1 - \lambda_u - \lambda_s)\mathcal{L}_{\text{cls}}^c + \epsilon H (\overline{\mathbf{p}}).
\end{equation}
We also found that compared to using voting methods based on classification results from a single modality, using a fusion modality, where features from two modalities are concatenated and then passed through a linear layer, can better capture the interaction information between modalities, thereby achieving better classification performance. Detailed comparison is shown in Appendix. The loss mentioned below is the same as previously described, except that the features are replaced with the new fusion features:
\begin{equation}
\mathcal{L}_{\text{fusion}} = \lambda_u (\mathcal{L}^u_{\text{rep}} + \mathcal{L}^u_{\text{cls}} ) + \lambda_s (\mathcal{L}^s_{\text{rep}} + \mathcal{L}^s_{\text{cls}} ).
\end{equation}
And, the overall objective for multimodal GCD learning is:
\begin{equation}
\mathcal{L} = \mathcal{L}_{\text{rep}} + \mathcal{L}_{\text{cls}}\label{unimodal_loss} + \mathcal{L}_{\text{fusion}}.
\end{equation}

\section{Results}
\label{sec:results}
\begin{table}[!tb]

\vspace{-1em}
\small  
\setlength{\tabcolsep}{4pt} 
\renewcommand{\arraystretch}{1.2} 
\begin{tabular}{m{0.15\linewidth}<{\centering}m{0.28\linewidth}<{\centering}m{0.32\linewidth}<{\centering}m{0.1\linewidth}<{\centering}}
\toprule
\centering\arraybackslash Dataset & \centering\arraybackslash Image & \centering\arraybackslash Text & \centering\arraybackslash Label \\ 
\midrule
\centering\arraybackslash N24News &  \centering\includegraphics[height=12mm, width=22mm]{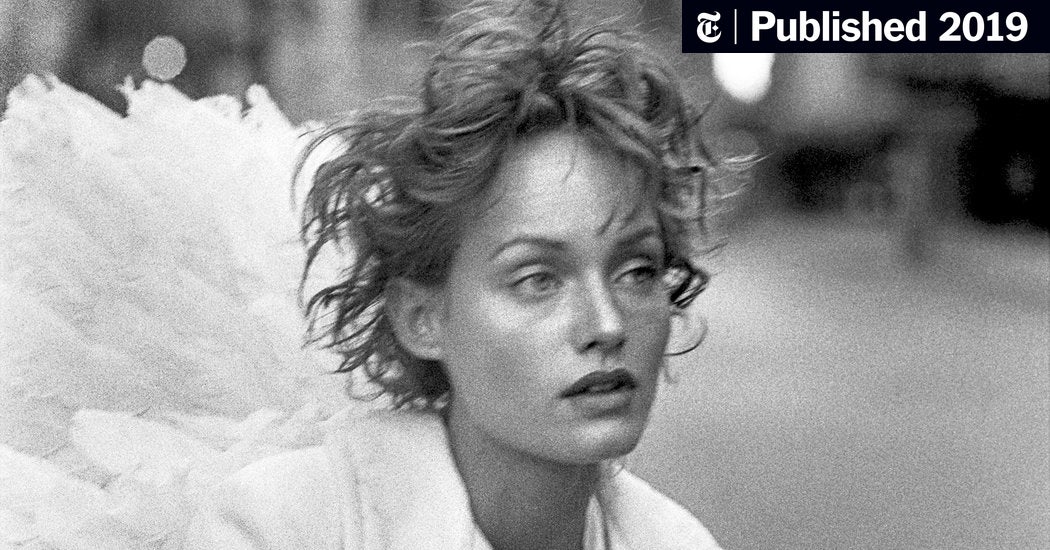} & \centering\arraybackslash Ms. Goodman styled Amber Valletta with wings... & Style\\
\centering\arraybackslash N24News &  \centering\arraybackslash\includegraphics[height=12mm, width=22mm]{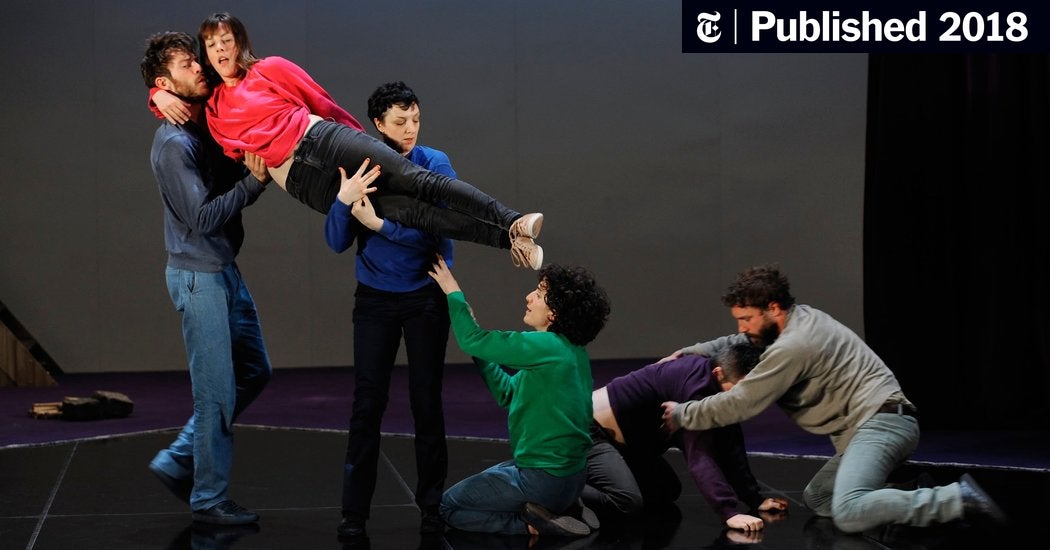} & \centering\arraybackslash A scene from Meg Stuart's "Until Our Hearts Stop"... &  Dance\\
\centering\arraybackslash Food101 &  \centering\arraybackslash\includegraphics[height=12mm, width=22mm]{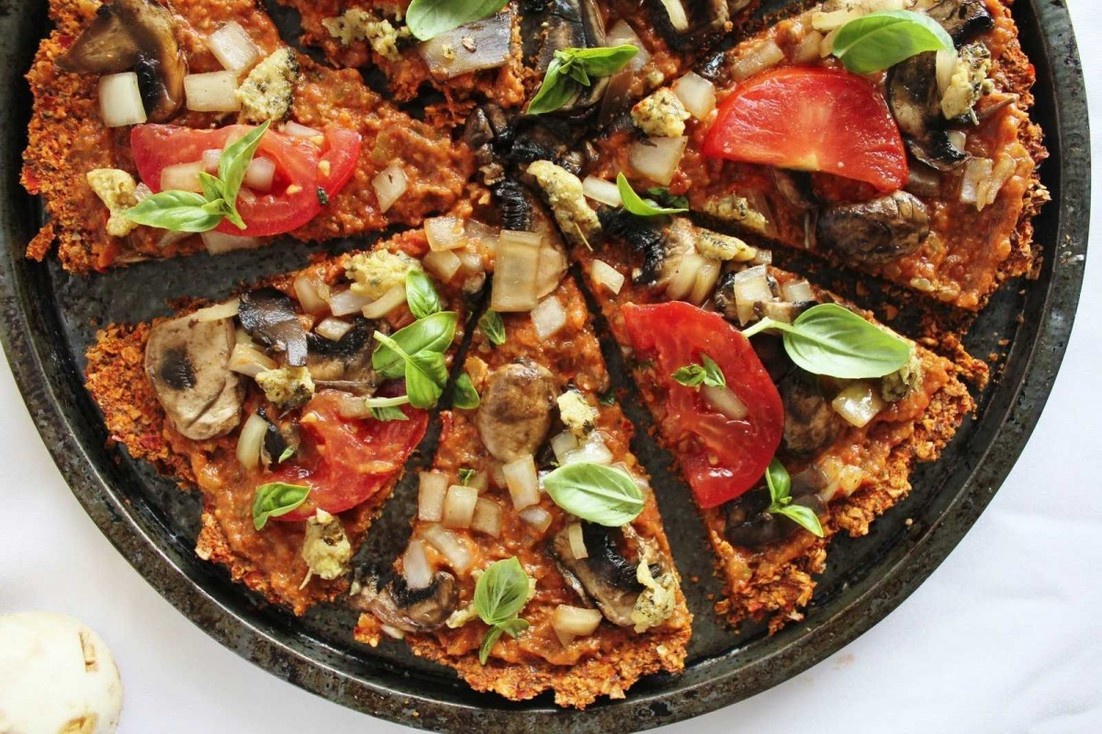} & \centering\arraybackslash Mind Blowingly Awesome Vegan Pizza & Pizza\\
\centering\arraybackslash Food101 &  \centering\arraybackslash\includegraphics[height=12mm, width=22mm]{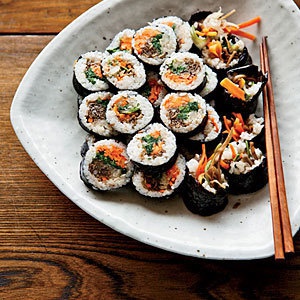} & \centering\arraybackslash Korean Sushi Rolls with Walnut-Edamame Crumble & Sushi\\
\bottomrule
\end{tabular}
\vspace{1em}
\caption{Data examples for each dataset.}
\label{tb: Datasets}
\vspace{-1em}
\end{table}

\subsection{Experiments Setup}
\textbf{Datasets. } 
We evaluate our approach on two image-text datasets: UPMC-Food-101 \cite{wang2015recipe} and N24News \cite{wang-etal-2022-n24news}. UPMC-Food-101 includes recipe descriptions paired with food images across 101 categories. N24News contains news images and four textual components, from which we use the Abstracts for pairing.  In both datasets, we selected 50\% of the categories as old classes, and from those, we picked out 50\% of the samples to form \( \mathcal{D}^l \), with the remainder serving as \(\mathcal{D}^u \). The specific divisions are detailed in Table~\ref{tab:split}.
\\\textbf{Evaluation Protocol. } 
We adopted the evaluation system from the original GCD work \cite{Vaze_2022_CVPR}. For the unlabeled data during training, classification accuracy is calculated by comparing the model-predicted labels with the true labels. The optimal match between predicted and ground truth labels is determined using the Hungarian algorithm \cite{kuhn1955hungarian}. In addition to overall accuracy on the entire $\mathcal{D}^u$ set, we also measured accuracy separately for old and new classes, referred to as \textbf{All}, \textbf{Old}, and \textbf{New} accuracy.
\begin{equation}
ACC = \max_{p \in P(\mathcal{Y}^u)} \frac{1}{M} \sum_{i=1}^{M} {1}(y_i = p(\hat{y}_i))
\end{equation}
\\
\noindent \textbf{Implementation Details.}
To support multimodal representations, we adopted the popular CLIP \cite{pmlr-v139-radford21a} model with ViT B/16 as the backbone. It is worth noting that, previous GCD papers \cite{Vaze_2022_CVPR,Wen_2023_ICCV} used ViT B/16 \cite{dosovitskiy2020image} pre-trained with DINO \cite{Caron_2021_ICCV} over ImageNet \cite{deng2009imagenet} as the backbone. To ensure a fair algorithmic comparison, we replicated those baselines with the CLIP backbone. 
We fine-tune the last transformer block of image and text encoder as well as the projection head. We employed a learning rate initially set to 0.1, with a cosine schedule, a batch size of 128, and trained for 200 epochs. Additionally, we adopted the practice from preivous work of setting the supervised weight to 0.35.

\begin{table}[!tb]

\vspace{-1em}
\begin{center}
\setlength{\tabcolsep}{4pt} 
\renewcommand{\arraystretch}{0.9} 
\resizebox{0.7\textwidth}{!}{%
\begin{tabular}{lcccccccc}
\toprule
& & &  \multicolumn{3}{c}{Food101} & \multicolumn{3}{c}{N24News}\\
\cmidrule(rl){4-6} \cmidrule(rl){7-9}
Methods & Sup  & Modal   & All  & Old  & New  & All  & Old  & New  \\
\midrule
\emph{MMBT} \cite{kiela2020supervisedmultimodalbitransformersclassifying} & \ding{51} & Fusion & \emph{92.1} & - & - & - & - & - \\ 
\emph{CMA-CLIP} \cite{liu2021cma} & \ding{51} & Fusion & \emph{93.1} & - & - & - & - & - \\ 
\emph{UniS-MMC} \cite{zou2023unismmc}  & \ding{51} & Fusion & \emph{94.7} & - & - & 84.7 & - & - \\ 
\midrule
\multirow{3}{*}{GCD~\cite{Vaze_2022_CVPR}} & \multirow{3}{*}{\ding{55}} & Visual & 40.0 & 68.7 & 25.3 & 27.8 & 30.1 & 27.0  \\ 
& & Text & 68.7 & 79.5 & 63.2 & 33.1 & 43.3 & 26.9 \\
& & Fusion & 61.7 & 75.5 & 54.8 & 32.8 & 44.2 & 26.0 \\
\midrule
\multirow{3}{*}{SimGCD~\cite{Wen_2023_ICCV}}  & \multirow{3}{*}{\ding{55}} & Visual & 65.5 & 75.0 & 60.8 & 44.7 & 54.0 & 39.2  \\ 
& & Text & 80.8 & 84.3 & 79.1 & 62.0 & \textbf{75.2} & 54.0 \\
& & Fusion & 73.1 & 85.7 & 66.8 & 63.8 & 75.1 & 57.1 \\
\midrule

{SPTNet~\cite{wang2024sptnet}}  & {\ding{55}} & Visual & 64.5& 71.4 & 61.0 & 44.1 & 53.5 & 38.5  \\ 
\midrule

{DPN ~\cite{an2023generalizedcategorydiscoverydecoupled}}  & {\ding{55}} & Text & 75.4 & 81.3 & 69.7 & 59.5 & 69.7 & 50.7  \\ 
\midrule
\multirow{3}{*}{Ours}                    & \multirow{3}{*}{\ding{55}} & Visual & 85.5 & 92.5 & 82.0 & 64.8 & 69.2 & 62.6  \\ 
& & Text & 85.3 & 92.3 & 81.9  & 65.1 & 68.9 & 62.8 \\
& & Fusion & \textbf{92.3} & \textbf{92.9} & \textbf{92.0} & \textbf{68.5} & 74.8 & \textbf{64.5} \\
\bottomrule
\end{tabular}%
}
\vspace{0.5em}
\caption{Results on the UPMC-Food101 Dataset~\cite{wang2015recipe} and N24News Dataset \cite{wang-etal-2022-n24news}.}
\label{tab:main_result}
\vspace{-3em}
\end{center}
\end{table}
\subsection{Main Results}
Due to the lack of publicly available multimodal semi-supervised models, we compared the state-of-the-art methods in current unimodal tasks and extended two of them (GCD \cite{Vaze_2022_CVPR} and SimGCD \cite{Wen_2023_ICCV}) to the multimodal setting. Specifically, we used CLIP's dual encoders to extract image and text features separately, and then concatenated these features to obtain fusion results, as described in Section 2. Since we have three different classifiers for image, text and fusion modality, we recorded the accuracy of the predicted labels from each classifier separately, denoted as \textbf{Visual}, \textbf{Text}, and \textbf{Fusion} in the table.

Additionally, we included fully supervised models like MMBT \cite{kiela2020supervisedmultimodalbitransformersclassifying}, CMA-CLIP \cite{liu2021cma}, and UniS-MMC
 \cite{zou2023unismmc} as upper-bound references, which were trained under full supervision without distinguishing between new and old categories.

\begin{itemize}
\item Our approach consistently improves performance by effectively aligning and harmonizing visual and textual features, unlike SimGCD where modality fusion can sometimes reduce accuracy.
\item The model shows strong capability in recognizing new categories, with cross-modal feature interaction enhancing robustness and reducing noise, leading to better novel category discovery.
\item On Food101, our fusion-based method outperforms fully supervised models like MMBT, even with only 50\% of classes labeled, demonstrating the potential for improving annotation efficiency in multimodal tasks without losing accuracy.
\end{itemize}

\subsection{Ablation Study}

\begin{figure}[!tb]
  \centering
  \hspace*{-0.07\columnwidth} 
  \includegraphics[width=1.1\columnwidth]{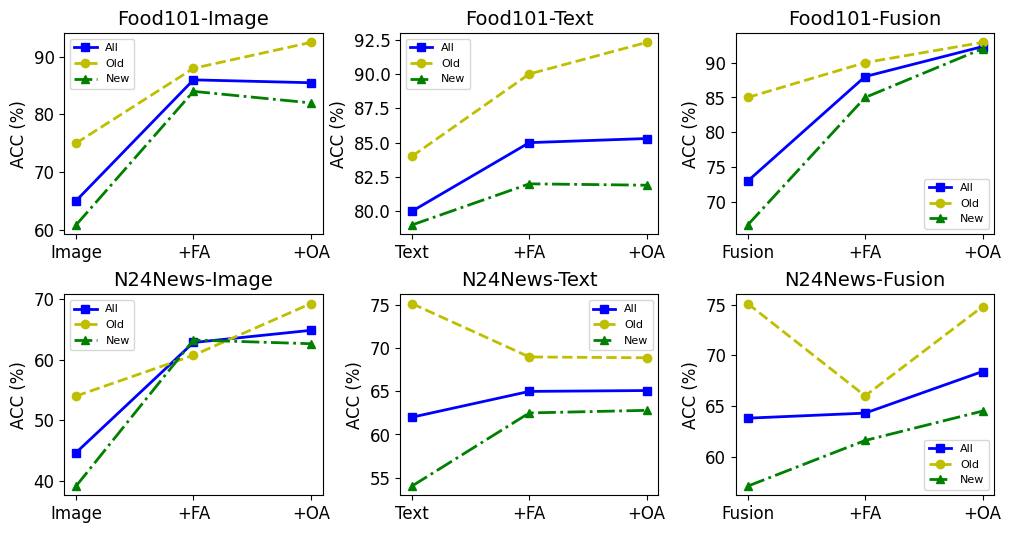}
  \vspace{-2em}
  \caption{Ablation study for different form of alignment. FA: add feature space alignment using multimodal contrastive learning. OA: add output space alignment using cross-modal distillation}
  \vspace{-1em}
  \label{fig:ablation}
\end{figure}
\begin{figure*}[!tb]
  \centering
  \includegraphics[width=0.95\textwidth]{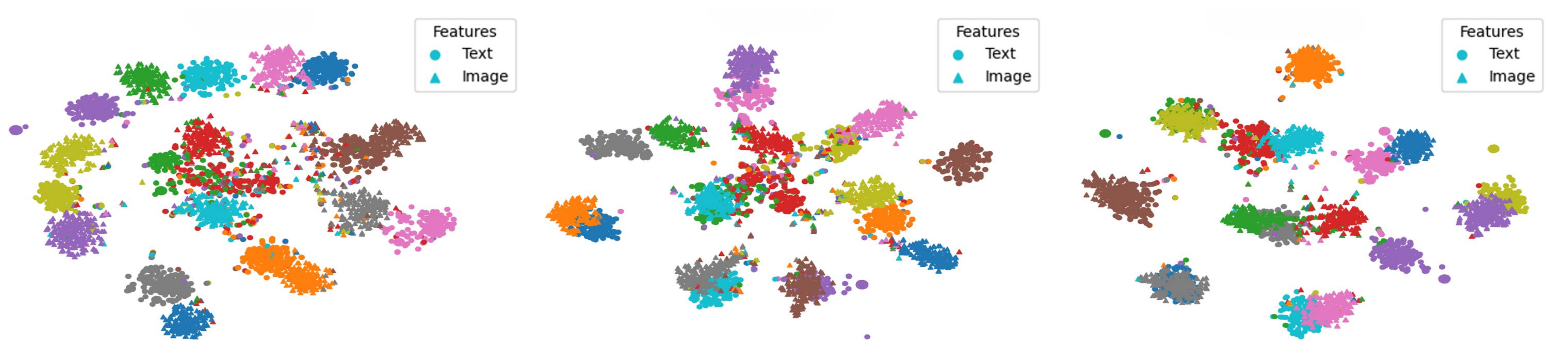} 
  \vspace{-1em}
  \caption{t-SNE result in Food101. From left to right: (\textbf{a}) Using the baseline methods, present unimodal GCD loss with image, text, and fused features; (\textbf{b}) Incorporating multimodal contrastive learning (feature align); (\textbf{c}) Adding multimodal prototype distillation (output align). Dots and triangles represent text and image respectively.}
  \vspace{-1em}
  \label{fig:t-SNE}
\end{figure*}

In this section, we provide a detailed step-by-step analysis of how our multimodal Generalized Category Discovery approach evolves from a straightforward baseline to the full model. Our approach utilizes three classifiers, each corresponding to \textbf{Image}, \textbf{Text}, and \textbf{Fusion} accuracy, respectively. Therefore, we report the accuracy changes for all three modalities at each step of the process. While we track the accuracy for image and text separately, our primary focus is on the fusion accuracy, as it reflects the combined effectiveness of the multimodal integration.

We start with the baseline method, where the SimGCD \cite{Wen_2023_ICCV} is applied separately to the image, text, and concatenated fusion features. This baseline setup provides the initial classification accuracy for each modality and serves as the foundation for subsequent enhancements.

The first enhancement introduces feature space alignment via cross-modal contrastive learning, as shown in the "FA" stage of Figure \ref{fig:ablation}. This step significantly improves accuracy, especially for "New" categories, by better leveraging complementary information across modalities. Slight drops in "Old" category accuracy on N24News are likely due to previous ways overfitting to well-represented old classes. We mitigate this by using rich infomation across different modalities. Details under unbalanced conditions are provided in Appendix A.

The final step involves output space alignment through an entropy-based loss, indicated by the "OA" stage. Even with minimal gains in individual modalities, the fusion results show substantial improvements. This confirms our theoretical insight that aligning predicted distributions across modalities creates a more coherent feature space, leading to the best performance when both alignments are applied.

Overall, aligning both feature and output spaces consistently boosts performance across categories, ensuring robust multimodal data handling for diverse and novel instances.

\section{Conclusion}
In conclusion, our research introduces a novel Multimodal Generalized Category Discovery (MM-GCD) framework, marking a significant advancement in handling mixed labeled and unlabeled datasets. By leveraging multimodal data, our approach not only addresses the limitations of traditional unimodal GCD but also demonstrates superior performance in identifying new categories across various benchmarks. This study sets a new standard for GCD applications, highlighting the potential of multimodal data in complex real-world classification tasks.

\bibliographystyle{unsrt}  
\bibliography{references}

\newpage
\section{Supplementary}
\subsection{Analysis of Prediction Bias}
\begin{figure}[h]
  \centering
  \hspace*{-0.07\columnwidth} 
  \includegraphics[width=0.8\columnwidth, height=0.5\columnwidth]{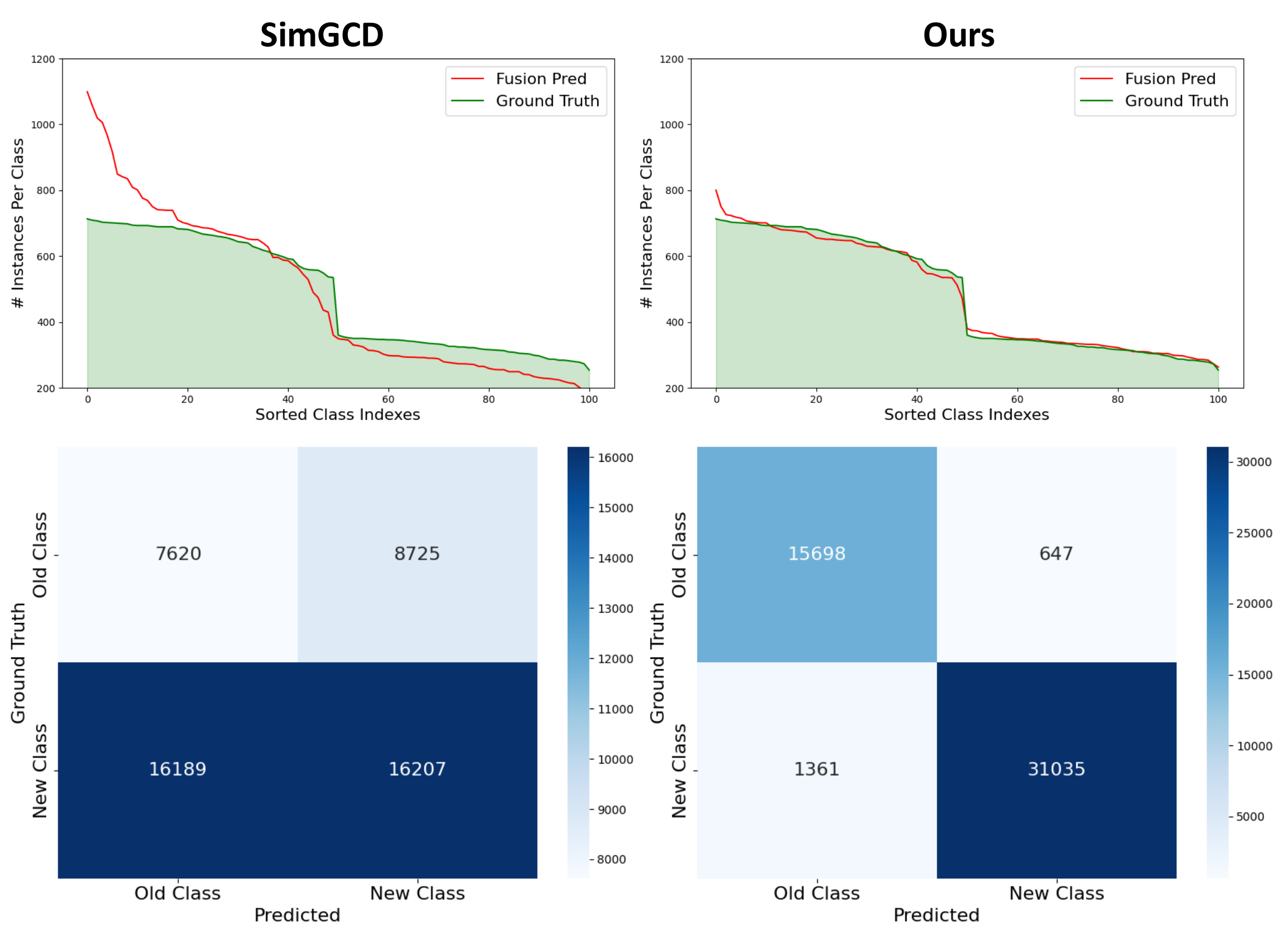}
  \caption{Comparison of prediction results distribution between SimGCD and our methods. The first row shows a comparison between the model's classification results and the ground truth at the category level. For clarity, all categories are sorted based on the number of ground truth samples, where classes 1-50 are old classes and classes 51-101 are new classes. The second row presents the confusion matrix for the binary classification of new vs. old classes. }
  \vspace{-10pt}
  \label{fig:balance-compare}
\end{figure}

We replicated the prediction results of SimGCD\cite{Wen_2023_ICCV} on the Food101 dataset\cite{wang2015recipe}, following the testing protocol where the ratio of new to old classes is set at 1:2. As shown in Fig.\ref{fig:balance-compare}, a comparison between SimGCD's predictions and the ground truth reveals a significant disparity. SimGCD tends to over-predict old classes and under-predict new ones, resulting in a pronounced bias towards old classes. The confusion matrix highlights this imbalance, which we believe is the primary factor behind SimGCD's lower accuracy on this dataset. The model's strong bias towards old classes limits its ability to generalize to novel categories.

In contrast, our model demonstrates a marked improvement in prediction balance, with predictions closely matching the true class distribution. The confusion matrix shows a more uniform distribution across both old and new classes, indicating reduced bias. We attribute this to our cross-modal supervision strategy, which enhances model robustness by integrating complementary information from different modalities. This effectively mitigates prediction imbalance, leading to more accurate and balanced classification, particularly in real-world scenarios with uneven class distributions.

\begin{table}[hbt]
\begin{center}
\resizebox{0.6\hsize}{!}{
\begin{tabular}{lcccccc}
\toprule
& \multicolumn{3}{c}{Food101} & \multicolumn{3}{c}{N24News} \\
\cmidrule(lr){2-4} \cmidrule(lr){5-7}
Methods & All & Old & New & All & Old & New \\
\midrule
Visual & 85.5 & 92.5 & 82.0 & 64.8 & 69.2 & 62.6 \\ 
Text & 85.3 & 92.3 & 81.9 & 65.1 & 68.9 & 62.8 \\
Voting & 87.5 & 90.5 & 85.9 & 67.8 & 72.1 & \textbf{65.3} \\
Fusion & \textbf{92.3} & \textbf{92.9} & \textbf{92.0} & \textbf{68.5} & \textbf{74.8} & 64.5 \\
\bottomrule
\end{tabular}
}
\vspace{5pt}
\caption{A comparison of different ways for final prediction.}
\label{tab:vote}
\vspace{-20pt}
\end{center}
\end{table}
\subsection{Comparison of Fusion Features and Soft Voting}
We compared the results of using soft voting, where the logits from text and image predictions are summed to select the final label, with those from using fusion features for prediction. We found that fusion features consistently outperformed soft voting. Detailed results are in table \ref{tab:vote}. We believe this is because fusion features more effectively integrate and refine semantic information from different modalities, offering richer context. In contrast, soft voting simply combines logits without fully capturing the correlations and complementarities between modalities, leading to reduced robustness and generalization on complex samples.

\subsection{Unknown Category Number}

Here we evaluate the GCD task performance with unknown category numbers. In previous experiments, we adopted the settings from previous work \cite{Vaze_2022_CVPR, Wen_2023_ICCV, zhang2023promptcal, wang2024sptnet}, which assumed that the number of new classes was known. This assumption can sometimes be difficult to meet in real-world scenarios. Therefore, it is necessary to first estimate the number of categories and then proceed with training based on the estimated number. 

\begin{table}[hbt]
  \begin{center}
  \begin{minipage}{0.4\textwidth} 
  \centering
  \resizebox{0.8\textwidth}{!}{
  \begin{tabular}{lcc}
  \toprule
     & Food101  & N24News \\
  \midrule
  Image  & 121 & 13 \\
  Text & 165 & 14 \\
  Fusion & \textbf{113} & \textbf{18} \\
  \midrule
  \emph{GT}  & \emph{101} & \emph{24} \\
  \bottomrule
  \end{tabular}
  }
  \vspace{-5pt}
  \subcaption{Estimated unknown class number.}
  \label{tab:estimate_num}
  \end{minipage}
  \hfill
  \begin{minipage}{0.48\textwidth} 
  \centering
  \resizebox{1\hsize}{!}{
  \begin{tabular}{lccccccc}
  \toprule
  &    & \multicolumn{3}{c}{Food101} & \multicolumn{3}{c}{N24News }\\
  \cmidrule(rl){3-5}\cmidrule(rl){6-8}
  Methods    &  $|\mathcal{C}|$  & All  & Old  & New  & All  & Old  & New \\
  \midrule
  SimGCD  & \emph{GT} & 73.1 & 85.7 & 66.8 & 63.8 & \textbf{75.1} & 57.1 \\
  Ours & \emph{GT} & \textbf{92.3} & 92.9 & \textbf{92.0} & \textbf{68.5} & 74.8 & \textbf{64.5} \\
  \midrule
  SimGCD & \emph{Estimated} & 75.6 & 75.7 & 75.5 & 60.9 & 75.0 & 52.4 \\
  Ours & \emph{Estimated} & 91.5 & \textbf{93.0} & 90.7 & 63.7 & 73.2 & 58.0 \\
  $\Delta$ & & \textcolor{darkgreen}{\textbf{+15.9}} & \textcolor{darkgreen}{\textbf{+17.3}} & \textcolor{darkgreen}{\textbf{+15.2}} & \textcolor{darkgreen}{\textbf{+2.8}} & \textcolor{red}{\textbf{-1.8}} & \textcolor{darkgreen}{\textbf{+5.6}} \\
  \bottomrule
  \end{tabular}
  }
  \vspace{-5pt}
  \subcaption{Comparison of SimGCD and our approach.}
  \label{tab:e_num_result}
  \end{minipage}
  \end{center}
  \vspace{-10pt}
  \caption{GCD task performance with unknown category numbers.}
  \vspace{-10pt}
  \end{table}

We estimate the number of new categories by following the off-the-shelf provided in original GCD \cite{Vaze_2022_CVPR}. Specifically, we utilize a pretrained model to generate corresponding features of a dataset without finetuning on that dataset. We then performe k-means clustering on these features with different numbers of categories $\textbf{K}$. The clustering result with the highest label accuracy is our estimated number of categories. As we are engaged in a multimodal classification task, we conducted this clustering for both vision and textual features, with the results presented in Table \ref{tab:estimate_num}. We found that a simple concatenation of features can significantly enhance the accuracy of initial category estimation. This demonstrates that multimodal information processed by multimodal models like CLIP can provide relatively more accurate clustering under initial conditions.

We then train the model using the number of categories esimated based on the fusion features. The results are shown in Table \ref{tab:e_num_result}. Our method consistently shows higher classification accuracy, demonstrating its robustness and generalization ability for real-world scenarios.


\subsection{More Visualization of Attention Maps}
In Fig.\ref{fig:attention_map} we present qualitative results to demonstrate the effects of our algorithm, especially compared with baseline algorithms including GCD and SimGCD. We adopt attention weights to generate visualized heatmaps, following common practices (\cite{wang2024get}\cite{zhang2023promptcal}) of interpreting Vision Transformer models. The visualization results are used to reflect the most essential spatial regions regarding their predictions. Showcases from the Food101 dataset are presented in Fig.\ref{fig:attention_map}. Compared with the baseline models, our model is shown to rely on comprehensive and discriminative features of the food itself, instead of noisy backgrounds. This indicates our model achieves not only more accurate predictions, but also a more interpretable and reliable decision process, which can be a critical advantage in specialized domains such as medicine and finance. 
\begin{figure}[hbt]
  \centering
  \includegraphics[width=0.8\columnwidth]{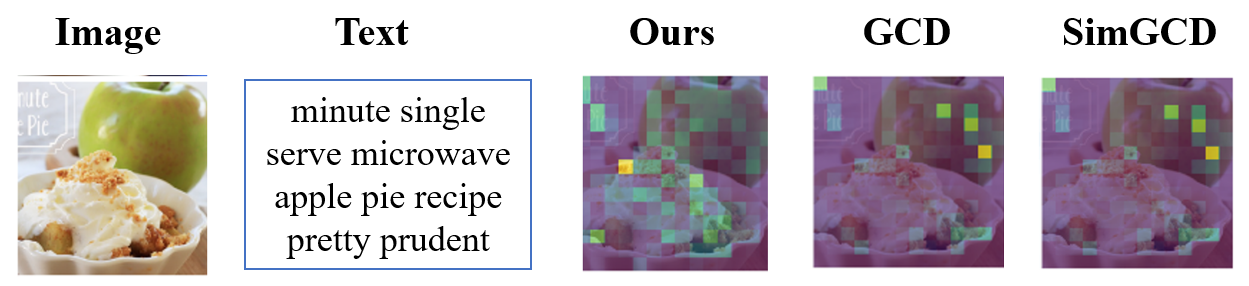}
  \vspace{-5pt}
  \caption{Sample of attention map visualization}
  \label{fig:apple_pie}
  \vspace{-20pt}
\end{figure}

Our method better utilizes the information from the textual side to help the image-side attention focus on the classification target. For example, Fig.\ref{fig:apple_pie} displays a picture of an apple pie. However, both SimGCD and the original GCD models only focus on the apples at the back of the image because they do not receive textual cues related to the pie. In contrast, our model effectively uses the information provided by the text, focusing more on the apple pie in the bowl. This demonstrates that when the images themselves are somewhat confusing or similar, textual information can effectively help the model focus on relevant content.

\subsection{More Visualization Comparison}
\begin{figure}[hbt]
  \begin{center}
    \begin{minipage}{0.45\textwidth} 
        \centering
        \includegraphics[width=0.8\columnwidth, height=6.4in]{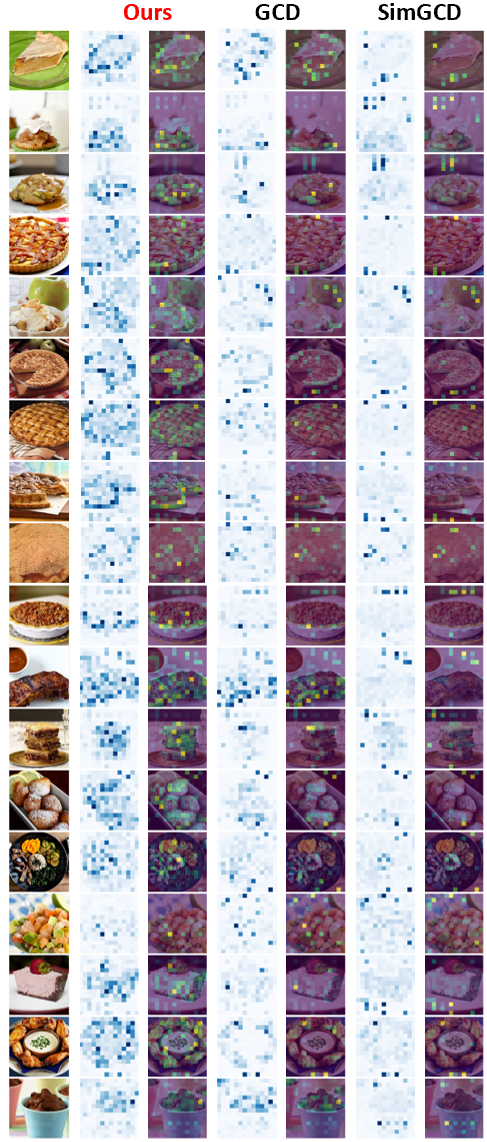}
        \captionof{figure}{Visualization of attention map}
        \label{fig:attention_map}
    \end{minipage}
    \hfill 
    \begin{minipage}{0.45\textwidth} 
        \vspace{-1em}
        \small  
        \setlength{\tabcolsep}{4pt} 
        \renewcommand{\arraystretch}{1.2} 
        \begin{tabular}{m{0.38\linewidth}<{\centering}m{0.1\linewidth}<{\centering}m{0.28\linewidth}<{\centering}}
        \hline
        \centering\arraybackslash Image & \centering\arraybackslash Label & \centering\arraybackslash Text \\  
        \hline
        \centering\includegraphics[height=12mm, width=22mm]{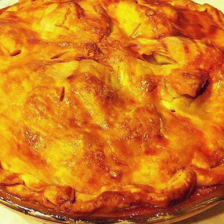} & Apple Pie & \centering\arraybackslash top thanksgiving recipes for every family dinner \\
        \hline
        \centering\includegraphics[height=12mm, width=22mm]{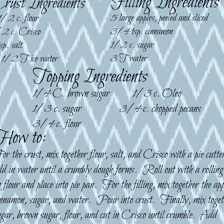} & Apple Pie & \centering\arraybackslash day homemade apple pie recipe imperfectly polished \\
        \hline
        \centering\includegraphics[height=12mm, width=22mm]{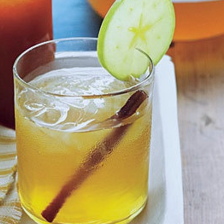} & Apple Pie & \centering\arraybackslash apple pie bourbon sweet tea recipe myrecipes com \\
        \hline
        \centering\includegraphics[height=12mm, width=22mm]{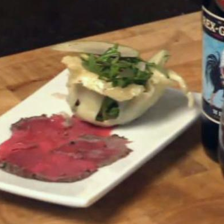} & Beef Carpaccio & \centering\arraybackslash beef carpaccio with apple and arugula salad recipe by chef billy parisi ifood tv \\
        \hline
        \centering\includegraphics[height=12mm, width=22mm]{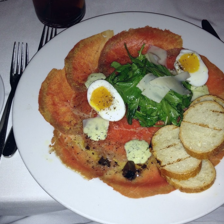} & Beef Carpaccio & \centering\arraybackslash kettler cuisine october \\
        \hline
        \centering\includegraphics[height=12mm, width=22mm]{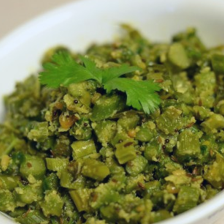} & Beef Carpaccio & \centering\arraybackslash lime marinated beef carpaccio with black olive tapenade and sun dried tomato pesto on foodrhythms \\
        \hline
        \centering\includegraphics[height=12mm, width=22mm]{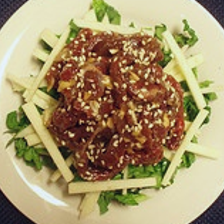} & Beef Tartare & \centering\arraybackslash the world most recently posted photos of yukhoe flickr hive mind \\
        \hline
        \centering\includegraphics[height=12mm, width=22mm]{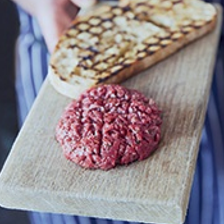} & Beef Tartare & \centering\arraybackslash pairing of the day october november \\
        \hline
        \centering\includegraphics[height=12mm, width=22mm]{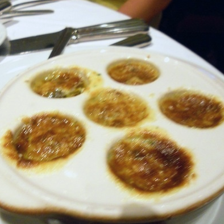} & Beef Tartare  & \centering\arraybackslash cruise dining much \\
        \hline
        \end{tabular}
        \captionof{table}{More visualization results on Food101. SimGCD predicts the category of these samples wrongly while our method predicts it correctly.}
        \label{tab:visual_compare}
        
    \end{minipage}
  \end{center}
\end{figure}

Table \ref{tab:visual_compare} presents additional visualization results from the Food101 dataset. The samples displayed below are cases where SimGCD made incorrect predictions, while our method correctly identified the categories. These examples involve significant noise in both text and images, making accurate classification challenging. The baseline approach struggles in such scenarios due to limited single-modality information processing. In contrast, our multimodal approach better integrates information across modalities, leading to more robust predictions. This highlights the advantages of leveraging complementary signals from both text and images in challenging conditions.



\end{document}